\documentclass[10pt,twocolumn,letterpaper]{article}

\usepackage{cvpr}
\usepackage{times}
\usepackage{epsfig}
\usepackage{graphicx}
\usepackage{amsmath}
\usepackage{amssymb}
\usepackage{hhline}
\usepackage{algorithm}
\usepackage[noend]{algpseudocode}
\usepackage{multirow}

\DeclareMathOperator*{\argmax}{arg\,max}

\DeclareMathOperator{\MSE}{MSE}

\usepackage[pagebackref=true,breaklinks=true,letterpaper=true,colorlinks,bookmarks=false]{hyperref}

\cvprfinalcopy 

\def\cvprPaperID{9883} 

\ifcvprfinal\fi
\begin{document}

\title{MCFlow: Monte Carlo Flow Models for Data Imputation}

\author{Trevor W. Richardson\\
\and
Wencheng Wu\\
\and
Lei Lin\\
\and
Beilei Xu\\
\and
Edgar A. Bernal\\
\and 
Rochester Data Science Consortium, University of Rochester\\ 
260 E.~Main St.~Suite 6108, Rochester, NY 14604 \\
{\tt\small {\{trevor.richardson, wencheng.wu, lei.lin, beilei.xu, edgar.bernal\}@rochester.edu}}
}\vspace{-1cm}

\maketitle

\begin{abstract}
We consider the topic of data imputation, a foundational task in machine learning that addresses issues with missing data.  To that end, we propose MCFlow, a deep framework for imputation that leverages normalizing flow generative models and Monte Carlo sampling.  We address the causality dilemma that arises when training models with incomplete data by introducing an iterative learning scheme which alternately updates the density estimate and the values of the missing entries in the training data.  We provide extensive empirical validation of the effectiveness of the proposed method on standard multivariate and image datasets, and benchmark its performance against state-of-the-art alternatives. We demonstrate that MCFlow is superior to competing methods in terms of the quality of the imputed data, as well as with regards to its ability to preserve the semantic structure of the data.
\end{abstract} \vspace{-0.6cm}

\section{Introduction}\vspace{-0.1cm}
Missing data is a widespread problem in real-life machine learning problems. Because most of the existing data analysis frameworks require complete datasets, imputation methods are indispensable to practitioners in the field.  As a consequence, data imputation has been the focus of extensive research in recent decades~\cite{little2002statistical,VANBUUREN2018}. Moreover, much of the recent research in data imputation has leveraged advanced machine learning techniques. This has resulted in the development of numerous shallow~\cite{10.1093/bioinformatics/17.6.520,mice2011, 10.1093/bioinformatics/btr597} and deep learning frameworks~\cite{pmlr-v32-rezende14,pmlr-v80-yoon18a,li2018learning,pmlr-v97-mattei19a} which have continuously pushed the state-of-the-art envelope.
A pervasive shortcoming of traditional learning-based imputation methods is that their training relies on fully observed data~\cite{PCM:1746339.1746348,pmlr-v32-rezende14}.  A more reasonable assumption, however, is that the training data itself may have missing entries, a limitation addressed by recently proposed methods~\cite{pmlr-v80-yoon18a,li2018learning,pmlr-v97-mattei19a}. In this paper, we propose a data imputation framework that leverages deep generative models and incorporates mechanisms that provide more accurate estimates of the missing data as compared to existing frameworks. In contrast to competing methods that are based on Generative Adversarial Networks (GANs)~\cite{pmlr-v80-yoon18a,li2018learning} and on Deep Latent Variable Models (DLVM)~\cite{pmlr-v97-mattei19a}, our framework leverages normalizing flow models~\cite{Dinh2014NICENI,45819,NIPS2018_8224}.  In particular, we take advantage of the exact log-likelihood evaluation, latent variable inference, and data sample reconstruction afforded by normalizing flow models in order to explicitly learn complex, high-dimensional data distributions.\vspace{-0.1cm}

We address the causality dilemma that arises when attempting to construct probabilistic models of incomplete data by adopting an alternating learning strategy inspired by model-based multiple imputation approaches~\cite{Little:1986:SAM:21412} such as those based on Expectation Maximization~\cite{Dempster77maximumlikelihood} and Monte Carlo Markov Chain~\cite{TakahashiMCMC} techniques.  Recent examples of related alternating techniques include applications to training generator networks~\cite{Han2016AlternatingBF,DBLP:conf/cvpr/GaoLZZW18}. The overarching idea behind the proposed framework is to alternately sample the data and update the density estimate until a good approximation of the true empirical distribution is attained.  This iterative process is enabled by the exact sampling and density evaluation aspects afforded by the normalizing flow model framework referenced above.  We note that although accurate density estimation is central to our framework, and autoregressive models achieve state-of-the-art performance in that field~\cite{VanDenOord:2016:PRN:3045390.3045575,pmlr-v80-chen18h,46840}, their slow sampling time would be a significance hindrance to our algorithm.
In order to address the problem of missing data, we introduce a new imputation algorithm called MCFlow. The proposed iterative learning scheme calls for alternately optimizing two different objective functions: (i) the traditional log-likelihood loss involved in training flow models, required to update the density estimate based on the complete data including imputed values; and, (ii) a maximum-likelihood criterion involved in sampling the latent flow space in order to find the optimal values of the missing data given the current estimate of the data distribution. While the former can be achieved by implementing known backpropagation techniques, we introduce a novel non-iterative approach to address the latter, which relies on a neural network trained to identify points in the latent space that maximize a density value while minimizing the reconstruction error as computed on the observed entries of the data. This approach can be viewed as an instance of an algorithm that learns to optimize~\cite{DBLP:journals/corr/LiM16b,DBLP:journals/corr/AndrychowiczDGH16} and, although not as effective as iterative~\cite{kelley1999iterative} or sampling-based methods~\cite{TakahashiMCMC,hastings70}, it is significantly more computationally efficient.

The primary contributions of this work can be summarized as follows:
\vspace{-0.3cm}
\begin{itemize}
	\setlength\itemsep{-0.2cm}
	\item a framework for data imputation based on deep generative normalizing flow models;
	\item an alternating learning algorithm that enables accurate density estimation of incomplete, complex and high-dimensional data by leveraging the efficient sampling and density evaluation attributes of flow models;
	\item a neural network that learns to optimize in the embedding flow space; and,
	\item extensive empirical validation of the proposed framework on standard multivariate and image datasets, including benchmarking against state-of-the-art imputation methods.
\end{itemize}\vspace{-0.3cm}
\section{Related Work}
An imputation method is classified as a single or multiple imputation method depending upon whether it estimates one or more than one value for each missing entry, respectively~\cite{little2002statistical}.  Even though a multitude of single imputation methods have been proposed in the literature~\cite{10.1093/bioinformatics/17.6.520,10.1093/bioinformatics/btr597,ATM8839}, multiple imputation methods are often preferred as they provide an assessment of uncertainty~\cite{little2002statistical,VANBUUREN2018,Murray2018MultipleIA}.  Multiple imputation frameworks often rely on building statistical models for the data, and then drawing samples from them in order to perform the imputation. Early attempts at building multiple imputation frameworks relied on simple parametric models such as Bayesian models~\cite{doi:10.1080/00949655.2015.1104683,doi:10.3102/1076998613480394} as well as mixtures of Gaussians~\cite{little2002statistical,DiZio:2007:ITF:1243526.1243912}.  More recently, with the advent of sophisticated deep generative models, the focus has shifted to investigating more effective ways to leverage the expressivity of the models to address data imputation scenarios.

Initial efforts exploiting deep models for imputation relied on the availability of fully observable training data~\cite{pmlr-v32-rezende14}.  More recent work has overcome this shortcoming.  Two such publications rely on modifications to the Generative Adversarial Network (GAN) architecture~\cite{NIPS2014_5423}.  The Generative Adversarial Imputation Network (GAIN) framework~\cite{pmlr-v80-yoon18a} employs an adversarially trained imputer optimized at discriminating between fake and real imputations.  While the GAN for Missing Data (MisGAN) approach~\cite{li2018learning} also implements an adversarially trained imputer, it additionally includes an explicit model for the missing data.  Unfortunately, being close GAN relatives, these models can be difficult to train~\cite{radford2015unsupervised,Arjovsky2017TowardsPM}. Furthermore, we believe there is an advantage to explicitly learning a probability density model of the data in imputation tasks, a feature that GAN-based frameworks generally don't afford.

Models that enable approximate density learning rely on estimating the variational bound~\cite{kingma2013autoencoding,pmlr-v32-rezende14,Burda2015ImportanceWA}.  While building such models requires observations of complete data, the Missing Data Importance-Weighted Autoencoder (MIWAE) framework~\cite{pmlr-v97-mattei19a} extends the variational lower bound principle to scenarios where only partial observations are available.  Unfortunately, such a family of methods is still intrinsically limited to learning an approximation of the density of the data and can therefore be challenging to optimize~\cite{NIPS2016_6581}.

In general, models that enable tractable density learning can be classified into frameworks based on fully visible belief networks (FVBNs) and those leveraging nonlinear independent components analysis (ICA).  FVBNs rely on the general product rule of probability which enables calculation of the joint distribution of a set of random variables using the product of conditional probabilities~\cite{Frey:1998:GMM:289988,oord2016wavenet}.  However, given the sequential nature of the operations involved, sampling the density estimate is not efficient. Nonlinear ICA methods define a set of continuous and invertible nonlinear mappings between two spaces~\cite{NIPS1994_901,Comon:2010:HBS:1841191,Dinh2014NICENI,45819,NIPS2018_8224}. Their main limitation stems from the fact that the mappings need to be invertible, which may limit their expressivity.\vspace{-0.1cm}

\section{Framework}\vspace{-0.1cm}
Throughout the paper, we consider the scenario in which data is Missing Completely at Random (MCAR)~\cite{Little:1986:SAM:21412}.  Formally speaking, assume that fully observable data points $x\in\mathcal{X}\subseteq\mathbb{R}^n$ are distributed according to $p_X(x)$, the set of binary masks $m\in\{0,1\}^n$ indicates the location of the missing data entries, and the mask entries are distributed according to $p_M(m|x)$. Data is MCAR when the missingness is independent of the data, that is, when $p_M(m|x)=p_M(m)$. 
Suppose that we observe data point $\tilde{x}^{(i)}$ with corresponding mask $m^{(i)}$, which means that the $k$-th entry in $\tilde{x}^{(i)}$, denoted $\tilde{x}^{(i)}_k$, is observed if $m^{(i)}_k=0$ and missing if $m^{(i)}_k=1$.  Often, it will be convenient to work with the complement of mask $m^{(i)}$, which we denote $\overline{m}^{(i)}$.  We use the notation $\tilde{x}^{(i)}_{\overline{m}^{(i)}}$ and $\tilde{x}^{(i)}_{m^{(i)}}$, respectively, to denote the set of observed and unobserved entries in $\tilde{x}^{(i)}$ given mask $m^{(i)}$.  Note that it is always possible to determine which entries of $\tilde{x}^{(i)}$ are missing and which are present, which means that $m^{(i)}$ can be uniquely determined from $\tilde{x}^{(i)}$.  Assuming $\tilde{x}^{(i)}$ is a partial observation of unknown data point $x^{(i)}$ with entries observed according to mask $m^{(i)}$, we formulate the data imputation task of recovering $x^{(i)}$ as a maximum likelihood problem, namely:\vspace{-0.4cm}

\begin{equation}
x^{(i)*}=\argmax\limits_{x^{(i)}} \left\{p_X(x^{(i)})\right\}\\
\text{s.t. } x^{(i)}_{\overline{m}^{(i)}}=\tilde{x}^{(i)}_{\overline{m}^{(i)}}
\label{eq:objective}
\end{equation}

Because $\log$ is a monotonic function, Eq.~\ref{eq:objective} is equivalent to\vspace{-0.4cm}

\begin{equation}
x^{(i)*}=\argmax\limits_{x^{(i)}} \left\{\log\left(p_X(x^{(i)})\right)\right\}\\
\text{s.t. } x^{(i)}_{\overline{m}^{(i)}}=\tilde{x}^{(i)}_{\overline{m}^{(i)}}
\label{eq:logobjective}
\end{equation}

Note that, in general, $p_X(\cdot)$ is unknown, but even if that were not the case, solving the optimization task from Eqs.~\ref{eq:objective} and \ref{eq:logobjective} would be challenging given that the distribution of any data of interest is likely to be highly non-convex and high-dimensional. One of the goals of this work is to make this optimization task feasible.  To that end, and for the sake of argument, assume a tractable, explicit density model for $p_X(x)$ exists.  Let this model be in the form of a generative network $G$ mapping a sample of interest $x \sim p_X(x)$ lying in space $\mathcal{X}$ to an embedding representation $z \sim p_Z(z)$ in space $\mathcal{Z}\subseteq\mathbb{R}^n$.  Further, assume that network $G$ effects a continuous, differentiable and invertible mapping $g:\mathcal{X}\rightarrow\mathcal{Z}$ such that $z = g(x)$ and 
\begin{equation}
p_X(x) = p_Z(g(x)) \left| \det \left({\frac{\partial g(x)}{\partial x^T}} \right) \right|
\label{eq:flow}
\end{equation}

This model is tractable if $p_Z(z)$ is tractable and if the determinant of the Jacobian of $g(\cdot)$ is tractable~\cite{DBLP:journals/corr/Goodfellow17}. Because $g(\cdot)$ is invertible, exact sample generation from density $p_X(x)$ is possible by drawing a sample $z \sim p_Z(z)$ and computing $x = g^{-1}(z)$. Additionally, computing the density on $x$ involves computing the density of its embedding vectors, and then scaling the result by the Jacobian determinant as in Eq.~\ref{eq:flow}~\cite{45819}.  Transformations that take the form of Eq.~\ref{eq:flow} fall under the category of nonlinear independent component analysis. A variety of methods of this type have been proposed in the
literature~\cite{Dinh2014NICENI,45819,NIPS2018_8224}.  In this paper, we implement network $G$ in the form of a normalizing flow model, and modify existing architectures to support learning from data with missing entries.  It is often computationally convenient to work with the log-likelihood function.  Taking the $\log$ of both sides of Eq.~\ref{eq:flow} results in\vspace{-0.5cm}

\begin{equation}
\log(p_X(x)) = \log(p_Z(g(x))) + \log\left(\left| \det \left({\frac{\partial g(x)}{\partial x^T}} \right) \right|\right)
\label{eq:logflow}
\end{equation}

Given a complete set of training data (that is, data without missing entries), and assuming $g(\cdot)$ is parameterized by a set of parameters $\theta$, learning network $G$ corresponds to finding an optimal set of parameters $\theta^*$ such that

\begin{equation}
\theta^* = \argmax\limits_{\theta} \left\{ \log(p_Z(g_{\theta}(x))) + \log\left(\left| \det \left({\frac{\partial g_{\theta}(x)}{\partial x^T}} \right) \right|\right) \right\}
\label{eq:flowtrain}
\end{equation}

If network $G$ were available, then the task of data imputation as formulated in Eq.~\ref{eq:logobjective} would become feasible.  For instance, standard gradient-based optimization techniques could be applied. Because missing data affects the training set in the scenarios under consideration, however, direct estimation of the distribution of the data according to Eqs.~\ref{eq:flow} or~\ref{eq:logflow} is not possible.

This scenario constitutes a causality dilemma.  Drawing inspiration from alternating algorithms such as Expectation Maximization (EM)~\cite{Dempster77maximumlikelihood}, MCFlow iteratively fills in the missing data according to Eq.~\ref{eq:logobjective} (i.e., based on the current estimate of the density), and updates the generative network parameters according to Eq.~\ref{eq:flowtrain} (i.e., based on the generated samples).  The intuition behind this approach is that finding the values of the missing entries in the data according to the maximum likelihood condition requires knowledge of the data distribution, and learning a model for the data distribution requires complete knowledge of the data.  As will become evident later, filling in for missing values in this iterative process involves drawing samples from the current model for the data distribution.  In this sense, the MCFlow framework is more closely related to Monte Carlo versions of the EM algorithm~\cite{doi:10.1080/01621459.1990.10474930,Neath2013OnCP} than to its vanilla version. In MCEM implementations, the E step consists in generating samples of the latent variables based on the current estimate of the predictive conditional distribution, and the M step estimates the parameters that maximize the observed posterior.  This similarity becomes more apparent if we interpret the imputation tasks from Eqs.~\ref{eq:objective} and \ref{eq:logobjective} as generating samples from the current approximation of the conditional predictive distribution of the missing data given the observed data and the current model parameters, $p_X(x|\theta,\tilde{x})$ (E step).  Furthermore, the optimization task from Eq.~\ref{eq:flowtrain} can be interpreted as updating the current approximation of the posterior of the model parameters given the observed and imputed values, $p(\theta|x,\tilde{x})$ (M step).

\section{Model Architecture}\label{sec:arch}

\begin{figure}[t]
	\centering
	\includegraphics[scale=0.25]{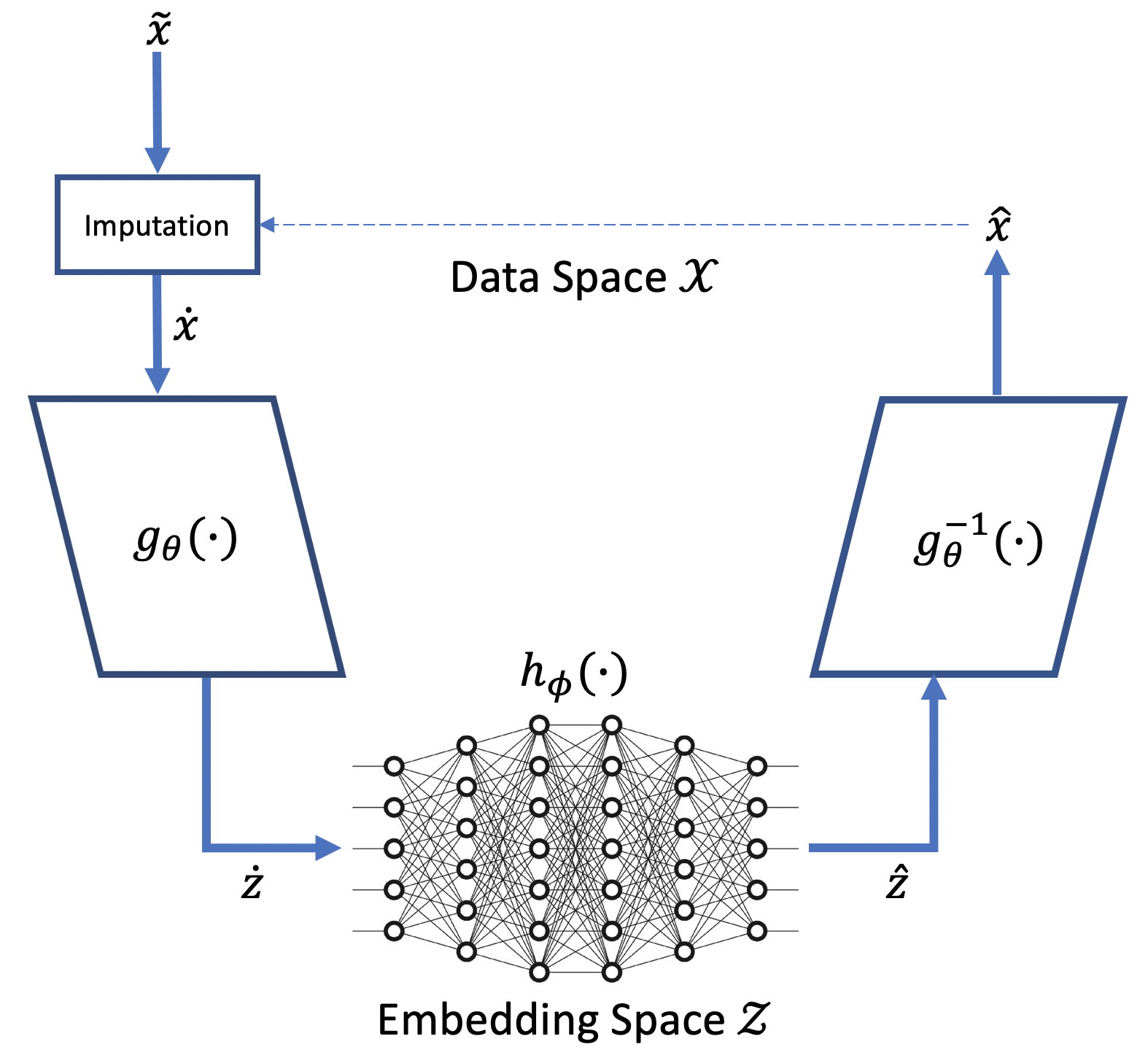}
	\caption{High-level view of the MCFlow architecture.} \label{fig:architecture} 
\end{figure}

The MCFlow architecture utilizes a hybrid framework comprised of a normalizing flow network $G$ (referred to herein as flow network or model) that is trained in an unsupervised manner, and a feedforward neural network that is trained in a supervised manner.  The flow network provides an invertible mapping $g_{\theta}(\cdot)$, where $\theta$ are the tunable parameters of the network, between data space $\mathcal{X}$ and embedding space $\mathcal{Z}$ and vice versa. The feedforward network $H$ operates in the embedding space by mapping input embedding vectors to output embedding vectors via function $h_{\phi}(\cdot)$, where $\phi$ are the tunable parameters of the network. In general, the role of the flow model is to learn the distribution of the data.  The role of the feedforward network is to find the embedding vector with the largest possible density estimate (i.e., the likeliest embedding vector) that maps to a data vector whose entries match the observed values (i.e., values at locations indexed by the complement of the mask). A high-level overview of the model is illustrated in Fig.~\ref{fig:architecture}.

\begin{figure*}
	\centering
	\includegraphics[scale=0.35]{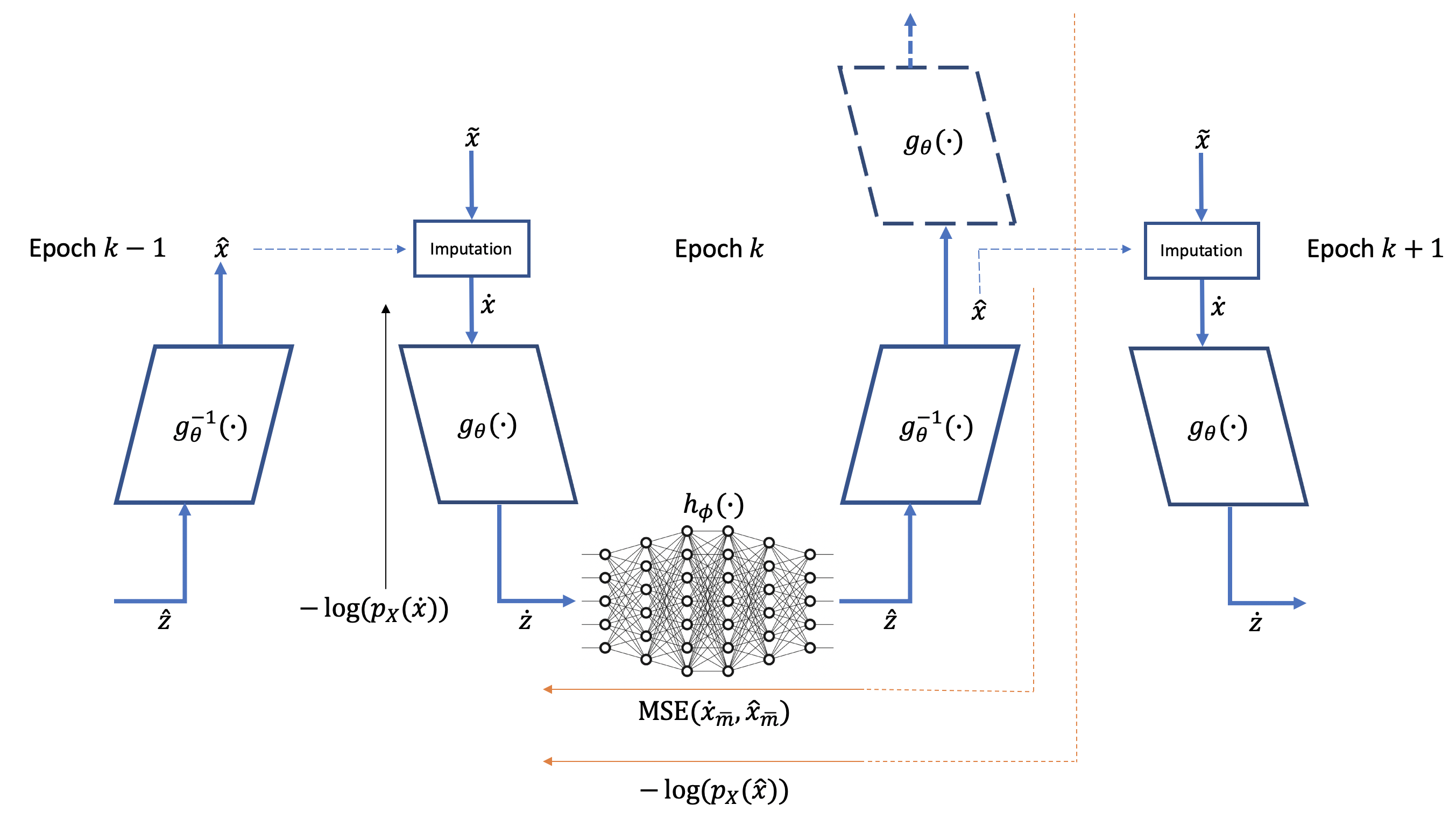}
	\caption{Unrolled view of the architecture and backpropagation process.} \label{fig:backprop} 
\end{figure*}

As with traditional flow implementations, training the generative portion of the framework involves finding the set of parameters $\theta^*$ that optimizes the mapping $g_{\theta}(\cdot)$ pursuant to Eq.~\ref{eq:flowtrain}. Because the optimization task from Eq.~\ref{eq:flowtrain} does not natively support incomplete data, a specific imputation scheme is implemented at initialization.  This scheme involves sampling the marginal observed density of the variable that contains the missing data in the case of multivariate, tabular data, and nearest-neighbor sampling in the case of image data.  After preprocessing the data as described above, an initial density estimate exists, and the model is now capable of performing data imputation. The estimates for missing values in $\tilde{x}$ are updated during defined subsequent iterations of training with the output of the model, $\hat{x}$. This is illustrated by the dotted arrow in Fig.~\ref{fig:architecture}. The alternating nature of the training process is more clearly illustrated in Fig.~\ref{fig:backprop}, which shows how the trained model from the previous epoch is used to update the missing value estimates for use in the current epoch.

The training stage of the generative portion of the framework can be formalized as follows: assume $N$ training samples, $\tilde{x}^{(i)}, i = 0, 1,\dots, N-1$ with corresponding masks $m^{(i)}$ are available.  For every incomplete sample $\tilde{x}^{(i)}$, a full training data sample $\dot{x}^{(i)}$ is computed by combining the observed values with imputed values from imputed sample $\hat{x}^{(i)}$ according to $\dot{x}^{(i)} = \tilde{x}^{(i)} \odot \overline{m}^{(i)} + \hat{x}^{(i)} \odot m^{(i)}$. Here, $\odot$ denotes the Hadamard, or element-wise product between two vectors.  With a full training set constructed, learning the optimal set of parameters $\theta^*$ in $g_{\theta}(\cdot)$ is accomplished by minimizing the following cost function:

\begin{equation}
-\frac{1}{N}\sum_{i = 0}^{N-1}  \log(p_X(\dot{x}^{(i)})) 
\label{eq:flowtrain0}
\end{equation}

\noindent or equivalently, using Eq.~\ref{eq:logflow}

\begin{equation}
-\frac{1}{N}\sum_{i = 0}^{N-1}  \left[\log(p_Z(g_{\theta}(\dot{x}^{(i)})) + \log\left(\left| \det \left({\frac{\partial g_{\theta}(\dot{x}^{(i)})}{\partial \dot{x}^{(i)T}}} \right) \right|\right)\right]
\label{eq:flowtrain1}
\end{equation}

\noindent which is the batch version of Eq.~\ref{eq:flowtrain}. In other words, the optimal parameters of network $G$ are the ones that maximize the log-likelihood of the imputed data, with the imputation mechanism being iteratively updated.  The black arrow in Fig.~\ref{fig:backprop} illustrates the direction in which the gradients are backpropagated when training the generative model.

Once an initial density estimate is available, learning the optimal mapping function $h_{\phi}(\cdot)$ in the embedding space is performed by training feedforward network $H$. The input to $H$ is the embedding of the imputed training set, namely $\dot{z}^{(i)}, i = 0, 1,\dots, N-1$, where $\dot{z}^{(i)} = g_{\theta}(\dot{x}^{(i)})$. Feedforward network $H$, which maps inputs $\dot{z}^{(i)}$ to outputs $\hat{z}^{(i)}$, is trained by finding the set of parameters $\phi$ in $h_{\phi}(\cdot)$ that minimizes the following objective function:\vspace{-0.5cm}

\begin{equation}
\frac{1}{N}\sum_{i = 0}^{N-1} \left[\MSE(\dot{x}^{(i)}_{\overline{m}^{(i)}},\hat{x}^{(i)}_{\overline{m}^{(i)}}) - \lambda\log(p_X(\hat{x}^{(i)})) \right]
\label{eq:nntrain}
\end{equation}

\noindent where $\dot{x}^{(i)} = g_{\theta}^{-1}(\dot{z}^{(i)})$, $\hat{x}^{(i)}= g_{\theta}^{-1}(\hat{z}^{(i)})$, $\hat{z}^{(i)}= h_{\phi}(\dot{z}^{(i)})$, and $\MSE(x,y)$ denotes the mean-squared error operator between vectors $x$ and $y$.  The first cost term in Eq.~\ref{eq:nntrain} encourages network $H$ to output an embedding whose reconstruction $g_{\theta}^{-1}(\hat{z}^{(i)})$ matches the training sample $\dot{x}^{(i)}$ at the observed entries.  The second cost term encourages network $H$ to output the vector with the highest density value according to the current density estimate.  Both terms combined yield estimates in the form of the likeliest embedding vector that matches the observed values, effectively solving the maximum likelihood objective from Eqs.~\ref{eq:objective} and~\ref{eq:logobjective}.  The red arrows in Fig.~\ref{fig:backprop} illustrate the computation of the different terms in Eq.~\ref{eq:nntrain}. The solid segments of the arrows indicate the sections of the framework that contain weights affected by the computations, while the dotted segments indicate where the computations take place and how they are propagated through the framework without affecting any parameters.  More specifically, the $\MSE$ term is computed in the data space but it only affects weights in neural network $H$ as it is backpropagated.  On the other hand, the log-likelihood term is computed in the embedding space of the flow model $G$ and is used to update the weights in $H$.

We note that optimizing the cost function from Eq.~\ref{eq:nntrain} requires repeated log-likelihood evaluation (e.g., in the computation of $\log(p_X(\hat{x}))$), latent variable inference (e.g., in the computation of $\dot{z}$ from $\dot{x}$), and data sample reconstruction (e.g., in the computation of $\hat{x}$ from $\hat{z}$), all of which can be effectively computed with flow network $G$.  While other generative models may outperform flow approaches at one task or another, we found flow models to constitute the best compromise for the needs of our framework.  Pseudocode for the training procedure is provided in Algorithm~\ref{code:training}.  Note that we concurrently update the parameters of $G$ and $H$ in order to achieve some computational savings: updating $G$ first and then $H$ would require two separate forward passes through $G$ (one to update $G$ itself and one to compute the mappings involved in updating $H$).

\begin{algorithm}
	\caption{Training Procedure}\label{code:training}
	\begin{algorithmic}
		\State \textbf{Data:} $N$ data points $\tilde{x}^{(i)}, i = 0, 1,\dots, N-1$ with corresponding masks $m^{(i)}$ indicating location of observed and missing entries
		\State \emph{Naively impute missing data in $\tilde{x}$ to produce $\dot{x}$:}
		\State $\dot{x} \leftarrow \tilde{x} \odot \overline{m} + \hat{x} \odot m$
		\For{$n=1$ to $nEpochs$}
		        \State \emph{Forward Pass:}
		        \State $\dot{z} \leftarrow g_{\theta}(\dot{x})$
		        \State $\hat{z} \leftarrow h_{\phi}(\dot{z})$
		        \State $\hat{x} \leftarrow g^{-1}_{\theta}(\hat{z})$
		        \State $\hat{z} \leftarrow g_{\theta}(\hat{x})$
		        \State \emph{Backpropagation:}
		        \State Compute loss according to Eq.~\ref{eq:flowtrain}
		        \State Compute loss according to Eq.~\ref{eq:nntrain}
		        \State Update $\theta$ and $\phi$ by backpropagating loss
		        \State $\dot{x} \leftarrow \tilde{x} \odot \overline{m} + \hat{x} \odot m$
		\EndFor
	\end{algorithmic}
\end{algorithm}

Once the model is trained, MCFlow imputes missing values according to Algorithm~\ref{code:imputation}.  The procedure involves encoding the naively imputed data sample $\dot{x}$ into $\dot{z}$, forward propagating $\dot{z}$ through the feedforward neural network to obtain $\hat{z}$, and reconstructing data sample $\hat{x}$ by decoding $\hat{z}$.  The final imputed sample $\dot{x}$ is obtained by filling in the missing entries with corresponding entries in $\hat{x}$
\begin{algorithm}
	\caption{Imputation Procedure}\label{code:imputation}
	\begin{algorithmic}
		\State \textbf{Data:} Test data point $\tilde{x}$ with corresponding mask $m$ 
		\State \emph{Naively impute missing data in $\tilde{x}$ to produce $\dot{x}$:}
		\State $\dot{x} \leftarrow \tilde{x} \odot \overline{m} + \hat{x} \odot m$
		    \State $\dot{z} \leftarrow g_{\theta}(\dot{x})$
		    \State $\hat{z} \leftarrow h(\dot{z})$ 
		    \State $\hat{x} \leftarrow g^{-1}_{\theta}(\hat{z})$
		    \State $\dot{x} \leftarrow \tilde{x} \odot \overline{m} + \hat{x} \odot m$
	\end{algorithmic}
\end{algorithm}\vspace{-0.5cm}

\subsection{Implementation Details}\label{sec:arch-imp-details}
The MCFlow architecture, comprised of various neural network layers, multiple optimizers and competing loss functions, can be implemented in the manner described in this section. A Pytorch implementation of the MCFlow algorithm is available online.\footnote{\href{https://github.com/trevor-richardson/MCFlow}{https://github.com/trevor-richardson/MCFlow}}
Preprocessing each dataset was done in two steps: 1) initializing $\tilde{x}^{(i)}_{m^{(i)}}$, and 2) scaling the data for training. Each data point $\tilde{x}^{(i)}$ requires the construction of initial estimates for $\tilde{x}^{(i)}_{m^{(i)}}$. This initial step can be performed easily using naive strategies such as zero imputation. We, however, employed two different initialization strategies, both of which estimate initial values for $\tilde{x}^{(i)}_{m^{(i)}}$ based on observable values in the dataset. For numerical multivariate datasets, each missing element in $\tilde{x}^{(i)}$ is replaced by sampling the marginal observed density of the variable containing the missing data. For the image datasets, each missing pixel was selected by randomly choosing one entry from the set of nearest observable neighbors of the missing pixel. After completing the selection of values for the missing data points, data was scaled to be in the interval $[0,1]$ via min-max normalization. 
For multivariate datasets, only observed values for each variable were used to determine the maximum and minimum values for that variable. For image datasets, the possible maximum of $255$ and minimum of $0$ were used for each pixel. These steps are required to construct the initial instantiation of $\dot{x}$.

\begin{table*}[ht]\caption{Imputation Results on UCI Datasets - RMSE (lower is better, 0.2 missing rate)} \label{tab:feature-table} 
\centering
\vspace*{5mm}
\begin{tabular}{|c|c|c|c|}
\hline
                     & Default Credit Card & Online News Popularity & Letter Recognition \\ \hline
MICE                 &  .1763       $\pm$    .0007        & .2585     $\pm$   .0010               &   .1537    $\pm$   .0006          \\ \hline
MissForest          &    .1623     $\pm$    .0120        &   .1976   $\pm$  .0015                &   .1605   $\pm$    .0004          \\ \hline
Matrix               &    .2282     $\pm$     .0005       &  .2602    $\pm$  .0073                &   .1442   $\pm$   .0006           \\ \hline
Auto-Encoder         &    .1667      $\pm$     .0014      &  .2388    $\pm$  .0005                &    .1351   $\pm$  .0009           \\ \hline
EM                   &     .1912   $\pm$    .0011         &   .2604  $\pm$     .0015              &      .1563 $\pm$       .0012      \\ \hline
GAIN                 & .1441 $\pm$ .0007             &  .1858 $\pm$ .0010     & .1198$\pm$ .0005   \\ \hline
\textbf{MCFlow}               & \textbf{.1233$\pm$.0012}  &  \textbf{.1760$\pm$.0032} & \textbf{.1033$\pm$.0017}  \\ \hline
\end{tabular}
\end{table*}

For all datasets and experiments, the Adam optimizer was used. For multivariate datasets, we used a learning rate of $1\times10^{-4}$ and a batch size of $128$. Normalizing flow network $G$ was built using affine coupling layers as introduced by the Real NVP framework~\cite{45819}. Our implementation of $G$ uses six affine coupling layers and a random masking strategy in contrast to the deterministic masking strategy used in Real NVP. The operations involved in the forward pass of each affine coupling transformation are depicted in Eq.~\ref{eq:affinecoupling}:
\begin{equation}
\begin{split}
y_{D}=x_{D},
\\
y_{\lnot D} = x_{\lnot D} \odot exp(s(x_{D})) + t(x_{D})
\end{split}
\label{eq:affinecoupling}
\end{equation}
In Eq.~\ref{eq:affinecoupling}, $D$ ($\lnot D$) represents the set of randomly selected indices which will not (will) be scaled or translated in the current affine transformation. Indices in $D$ were initialized using a binomial distribution with a $50\%$ success rate. These indices remain constant after initialization. In our implementation of $G$, the $s$ and $t$ functions of every coupling layer are defined by 4-layer fully connected neural networks. Both $s$ and $t$ networks use Leaky ReLu as the activation function in the hidden layers. The final output layer for $s$ and $t$ utilizes the activation functions $\tanh$ and linear, respectively.

Network $H$ has five linear layers with the same number of neurons in each layer as the dimensionality of the data. Leaky ReLu was chosen as the activation function between layers. For the image datasets, the following changes were made as a result of resource constraints: First, the batch sizes used for MNIST, CIFAR-10 and CelebA were $128$, $128$ and $512$ respectively; second, for missing data rates above $60\%$, a learning rate of $1\times10^{-3}$ was employed. A detailed description of the training procedure for MCFlow can be seen in Algorithm \ref{code:training}.

MCFlow periodically updates the missing entries in the training data and resets the $\theta$ parameters in the flow model function, $g_{\theta}(\cdot)$. To that end, we use an exponential scheduling mechanism, where the data update and parameter reset occurs at every epoch that is a power of 2. This implies that in order to perform inference on arbitrary data points, it is necessary to save the model parameters for both $h_{\phi}(\cdot)$ and $g_{\theta}(\cdot)$ after updating $\dot{x}$ according to Eq.~\ref{eq:resetdistr} but before resetting the model parameters in $g_{\theta}(\cdot)$:
\begin{equation}
\dot{x} = \tilde{x} \odot \overline{m} + \hat{x} \odot m
\label{eq:resetdistr}
\end{equation}

\begin{table*}[ht]\caption{Imputation Results on Image Datasets - RMSE (lower is better)} \label{tab:imaging-table} 
\centering
\setlength\doublerulesep{0.5pt}
\vspace*{5mm}
\begin{tabular}{|c|c|c|c|c|c|c|c|c|c|c|}
\hline
            &            Missing Rate $\rightarrow$                    & .1      & .2      & .3      & .4      & .5      & .6      & .7      & .8      & .9      \\ \hline
\multirow{3}{*}{MNIST}           & GAIN & .11508        &  .12441       &    .13988     &  .14745       &  .16281       & .18233        &  .20734       & .24179        & .27258 \\  
                                                  & MisGAN   & .11740        & .10997       &  .11377       & .11297        &  .12174       & .13393        &  .15445       & .19455        & .27806       \\  
                                                  & \textbf{MCFlow}   & \textbf{.07464} & \textbf{.07929} & \textbf{.08508} & \textbf{.09187} & \textbf{.10045} & \textbf{.11255} & \textbf{.12996} & \textbf{.15806} & \textbf{.20801}         \\ \hline
\multirow{3}{*}{CIFAR-10}                 & GAIN    &    .10053 & .12700       & .13248        &  .11785       &  .12451       &  .13130       &  .13832       &  .18728       & .53728 \\  
                                                 & MisGAN & .18923        &  .15223       &  .14746       & .12947        & .13027        & .14746        & .17335        & .24060        & .31722        \\ 
                                                 & \textbf{MCFlow}   & \textbf{.06012} & \textbf{.06232} & \textbf{.06686} & \textbf{.07215} & \textbf{.08311} & \textbf{.10048} & \textbf{.13132} & \textbf{.15015} & \textbf{.16939}        \\ \hline
\multirow{2}{*}{CelebA}    & GAIN & .06752  & .07493      &  .08367       & .08479        &  .09292       &  .10608       &  .11720       &  .14042       &   .52050            \\ 
                                                 
                                                  & \textbf{MCFlow}   & \textbf{.05733}       &  \textbf{.06243}       & \textbf{.06266}        & \textbf{.06946}        & \textbf{.07261}        & \textbf{.07890}        & \textbf{.08487}       &  \textbf{.11073}       &  \textbf{.12225}      \\ \hline
\end{tabular}
\end{table*}
\begin{table*}[h!]\caption{FID Results on Imputed MNIST Data (lower is better)} \label{tab:fid-table} 
\centering
\setlength\doublerulesep{0.5pt}
\vspace*{5mm}
\begin{tabular}{|c|c|c|c|c|c|c|c|c|c|}
\hline
 Missing Rate $\rightarrow$                         &      .1      & .2      & .3      & .4      & .5      & .6      & .7      & .8      & .9      \\ \hline
GAIN & .0696        &  .4035       &    1.24     &  3.277       &  6.337      & 12.44        &  22.91       & 43.69        & 92.74 \\  
\textbf{MisGAN}   & .0529       & .1015       &  \textbf{.2085}       & \textbf{.2691}        &  \textbf{.3634}       & \textbf{.8870}        &  \textbf{1.324}       & \textbf{2.334}        & \textbf{6.325}       \\  
MCFlow   & \textbf{.0521} & \textbf{.0779} & .2295 & .6097 & .8366 & .9082 & 1.951 & 6.765 & 15.11         \\ \hline

\end{tabular}
\end{table*}
\begin{table*}[h!]\caption{Classification Accuracy on Imputed MNIST Data (higher is better)} \label{tab:classification-table} 
\centering
\setlength\doublerulesep{0.5pt}
\vspace*{5mm}
\begin{tabular}{|c|c|c|c|c|c|c|c|c|c|}
\hline
 Missing Rate $\rightarrow$                         &      .1      & .2      & .3      & .4      & .5      & .6      & .7      & .8      & .9      \\ \hline
GAIN & .989        &  .988       &    .985     &  .978       &  .969       & .931        &  .852       & .629        & .261 \\  
MisGAN   & .989        & .988       &  .986       & .980        &  .968       & .945        &  .872       & .690        & .334       \\  
\textbf{MCFlow}   & \textbf{.991} & \textbf{.990} & \textbf{.990} & \textbf{.988} & \textbf{.985} & \textbf{.979} & \textbf{.963} & \textbf{.905} & \textbf{.705}         \\ \hline

\end{tabular}
\end{table*}

Based on this exponential scheduling scheme, inference in the MCFlow architecture, as described in Algorithm \ref{code:imputation}, requires all of the models saved during training. The number of saved models and passes through the architecture is a function of the number of epochs trained, $M$. More specifically, training MCFlow involves saving the parameters for $h_{\theta}(\cdot)$ and $g_{\theta}(\cdot)$ at every epoch that is a power of 2, ultimately requiring $\lceil\log_2(M)\rceil$ saved models in order to properly impute missing data for new samples. The longest the model took to converge was $500$ epochs, which required nine saved models to properly inference testing datapoints.  Imputing missing data in the test dataset involves performing initial naive imputation followed by full passes through each saved architecture. At the end of this process, the final prediction from MCFlow of $x$ in $\hat{x}$ is returned and performance metrics are recorded. 

\section{Experimental Results}\label{sec:results}


\subsection{Datasets}\label{sec:results-dataset}

We evaluate the performance of MCFlow as well as competing methods on three standard, multivariate datasets from the UCI repository \cite{Dua:2019} and three image datasets. The UCI datasets considered are Default of Credit Card Clients, Letter Recognition and Online News Popularity. The image datasets used in this research are MNIST\cite{lecun-mnisthandwrittendigit-2010}, CIFAR-10~\cite{liu2015faceattributes} and CelebA~\cite{krizhevsky2009learning}. The MNIST dataset contains $28\times28$-pixel grayscale images of handwritten digits; we used the standard 60,000/10,000 training/test set partition. CIFAR-10 contains $32\times32$-pixel RGB images from ten classes; we used the standard 50,000/10,000 training/test set partition. CelebA contains $178\times218$-pixel RGB images of celebrity faces. It was split using the first 162,770 images as the training set and the last 19,962 images for testing. The CelebA images were center cropped and resized to $32\times32$ pixels using bicubic interpolation. 

\subsection{Experimental Setup}

Imputation performance on multivariate and image data was measured using using root mean squared error (RMSE). We report performance numbers on each test set from models obtained upon convergence of the training process, where convergence is determined based on stabilization of the training losses for both $G$ (Eq.~\ref{eq:flowtrain1}) and $H$ (Eq.~\ref{eq:nntrain}). We emphasize that the training set itself has missing data and that the training loss is computed only on observed data points in order to mimic real-world scenarios as faithfully as possible. Imputation performance on the UCI datasets was evaluated with a data missingness rate of $20\%$ using five-fold cross validation. We report the mean and standard deviation of the imputation accuracy across all folds for six competing methods: MICE~\cite{mice2011}, MissForest~\cite{10.1093/bioinformatics/btr597}, Matrix Completion (Matrix)~\cite{Mazumder:2010:SRA:1756006.1859931}, Auto-Encoder~\cite{Gondara2017MultipleIU}, Expectation-Maximization (EM)~\cite{PCM:1746339.1746348} and GAIN~\cite{pmlr-v80-yoon18a}. For MNIST, CIFAR-10 and CelebA, the testing imputation accuracies are reported across missing rates ranging from $10\%$ to $90\%$ in steps of $10\%$. Competing methods considered include GAIN~\cite{pmlr-v80-yoon18a} and MisGAN~\cite{li2018learning}. All reported numbers are measured with respect to imputation accuracy over the missing data, namely $\tilde{x}^{(i)}_{m^{(i)}}$. Despite considerable efforts to make MisGAN achieve reasonable performance numbers on CelebA, we weren't able to manage it based on the currently published code. No other versions of the code were available directly from the authors. We also measured the quality of the imputed MNIST images with the Fréchet Inception Distance (FID)~\cite{heusel2017gans}, which has been shown to correlate well with human perception. Lastly, we measured the ability of the algorithms to preserve semantic content by performing a classification task on imputed imagery with networks pretrained on fully observed data.

\subsection{Quantitative Results}

Tables~\ref{tab:feature-table} and~\ref{tab:imaging-table} depict MCFlow's performance at MCAR imputation against competing methods. For all datasets and missing rates, the MCFlow framework outperforms all other methods at imputation accuracy from the perspective of the RMSE between the predictions of the model in question and the ground truth values. For the UCI datasets, MCFlow produces an average reduction in square error of $11\%$ as compared to the existing state of the art, GAIN. In addition, MCFlow produces an average reduction in RMSE of $19\%$, $38\%$ and $27\%$ as compared to the next best performing method on MNIST, CIFAR-10 and CelebA respectively.  Sec.~1 in the Supplementary Material contains imputation results on the training dataset and in terms of PSNR.  The results in Table~\ref{tab:imaging-table} illustrate the pixel-level quality of the imputed imagery; in contrast, Table~\ref{tab:fid-table} includes FID performance which is meant to be indicative of quality as perceived by humans. It can be seen that MisGAN outperforms competing methods across most of the missingness range.  As will be illustrated in Sec.~\ref{sec:qual}, while this means that imputed images more closely resemble attributes of the target image population, they do not necessarily preserve the original semantic content of the partially observed inputs.


In order to illustrate the performance of the methods beyond image quality metrics, we tested their imputation abilities within the context of a larger data processing pipeline where incomplete data is imputed before being processed through a classifier trained on complete data.  To that end, we measured the classification performance of a LeNET-based handwritten digit classifier on imputed MNIST data with various degrees of missingness.  The LeNET network was pre-trained on MNIST data without missing values.  Table~\ref{tab:classification-table} contains these results. It can be seen that the imputation results produced by MCFlow have the smallest impact on the semantic content of the imagery, as classification results are consistently higher for images imputed with our method.  This phenomenon becomes more evident as the missing data rate increases: the classifier operating on MCFlow-imputed imagery is able to achieve good classification accuracy up to the highest rate of missingness tested, and performs acceptably even in this extreme scenario.

\subsection{Qualitative Results}\label{sec:qual}
\begin{figure}[t]
	\centering
	\includegraphics[scale=0.85]{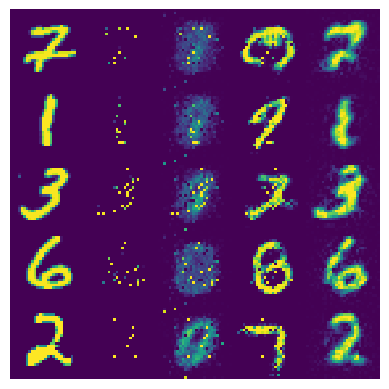}\\ \vspace{-0.3cm}
	(a) \hspace{0.36in} (b)\hspace{0.36in} (c) \hspace{0.36in} (d)\hspace{0.36in} (e)\\
	\caption{Sample imputation results for MNIST at 0.9 missingess rate: (a) Original, (b) observed, (c) GAIN-imputed, (d) MisGAN-imputed, and (e) MCFlow-imputed images.} \label{fig:mnistresults} 
\end{figure}

Fig.~\ref{fig:mnistresults} illustrates the imputation performance of competing methods for the $90\%$ missing data rate on the MNIST dataset. Images along columns (a) and (b) contain the original and observed images, respectively. Pixels that are not observed are assigned a value of 0 for visualization purposes. The imputation models only see the observed images; complete images are included for reference only.  Images along columns (c), (d) and (e) include the imputed results after using GAIN, MisGAN and MCFlow, respectively.  It can be seen that MCFlow does the best job among the competing methods at preserving and recovering the semantic content of the intended image, which further supports the results from Table~\ref{tab:classification-table}.  Numbers in GAIN-imputed images are, for the most part, illegible. In contrast, MisGAN-imputed images are visually impressive, which is in line with the results from Table~\ref{tab:fid-table}. One shortcoming of MisGAN, however, is that the imputations produced often fail to represent the digit contained in the original image. Sec.~2 in the Supplementary Material contains additional qualitative results on MNIST and CIFAR-10.  

\section{Conclusions}
We proposed MCFlow, a method for data imputation that leverages a normalizing flow model as the underlying density estimator. We augmented the traditional generative framework with an alternating learning scheme that enables it to accurately learn distributions from incomplete data sets with various degrees of missingness.  We empirically demonstrated the superiority of the proposed method relative to state-of-the-art alternatives in terms of RMSE between the imputed and the original data. Experimental results further show that MCFlow outperforms competing methods with regards to preservation of the semantic structure of the data. This is evidenced by the superior classification performance on imputed data achieved by a classifier trained on complete data, which holds across the full range of missing data ratios evaluated.  The ability of the method to make sense of and recover the semantic content of the data at every tested missing rate indicates that it is effectively learning the underlying statistical properties of the data, even in extreme paucity scenarios.

\small{\textbf{Acknowledgements.} The authors would like to thank Charles Venuto and Monica Javidnia from the Center for Health and Technology (CHET) at the University of Rochester for their fruitful insight and support.}


{\small
\bibliographystyle{ieee_fullname}
\bibliography{egbib}
}

\end{document}